\documentclass{article}
\usepackage{spconf,amsmath,graphicx,hyperref}
\usepackage{booktabs}
\usepackage{float}
\usepackage{wrapfig}
\usepackage{soul, xcolor}
\usepackage{enumitem}

\newcommand{\rev}[1]{\textcolor{black}{#1}}

\title{Rethinking Large Language Models for Irregular Time Series Classification in Critical Care}
\name{Feixiang Zheng$^{\ast}$, Yu Wu$^{\dagger}$, Cecilia Mascolo$^{\dagger}$, Ting Dang$^{\ast}$}
\address{$^{\ast}$The University of Melbourne, Australia \ $^{\dagger}$University of Cambridge, UK}
\begin{document}
\ninept
\maketitle
\begin{abstract}
Time series data from the Intensive Care Unit (ICU) provides critical information for patient monitoring. While recent advancements in applying Large Language Models (LLMs) to time series modeling (TSM) have shown great promise, their effectiveness on the irregular ICU data, characterized by particularly high rates of missing values, remains largely unexplored. This work investigates two key components underlying the success of LLMs for TSM: the \textit{time series encoder} and the \textit{multimodal alignment strategy}. To this end, we establish a systematic testbed to evaluate their impact across various state-of-the-art LLM-based methods on benchmark ICU datasets against strong supervised and self-supervised baselines. Results reveal that the encoder design is more critical than the alignment strategy. Encoders that explicitly model irregularity achieve substantial performance gains, yielding an average AUPRC increase of 12.8\% over the vanilla Transformer. While less impactful, the alignment strategy is also noteworthy, with the best-performing semantically rich, fusion-based strategy achieving a modest 2.9\% improvement over cross-attention. However, LLM-based methods require at least 10$\times$ longer training than the best-performing irregular supervised models, while delivering only comparable performance. They also underperform in data-scarce few-shot learning settings. These findings highlight both the promise and current limitations of LLMs for irregular ICU time series. The code is available at \href{https://github.com/mHealthUnimelb/LLMTS}{https://github.com/mHealthUnimelb/LLMTS}.

\end{abstract}

\vspace{-5pt}
\section{Introduction}
\vspace{-5pt}


\rev{The rich, multivariate time series data generated in the Intensive Care Unit (ICU), such as vital signs and lab tests \cite{zheng2025development, mTAN}, holds immense potential for improving patient outcomes through machine learning. However, this potential is fundamentally challenged by the nature of ICU data: it is profoundly irregular, characterized by sporadic measurements, high missingness rates, and asynchronous sampling across sensors, which makes effective modeling a significant hurdle.}

\rev{Prior research has explored irregular-aware data augmentation and time-aware encoding \cite{mTAN,latent_ODE,warpformer,primenet, StatioCL}. While effective on specific datasets, these models often fail to generalize robustly across different hospitals or patient populations, which may use different monitoring devices with sensor-specific dynamics or exhibit unique irregular patterns. The recent emergence of Large Language Models (LLMs) offers a compelling alternative, distinguished by their powerful zero-shot and few-shot generalization capabilities \cite{brown2020language, achiam2023gpt}. Early studies have successfully applied LLMs to time series~\cite{time_llm, S2IP_llm}, benefiting from their ability to create aligned embeddings of time series and unstructured text, thereby providing a richer source of information for modeling. However, this research has predominantly focused on clean, regularly-sampled time series, leaving a critical gap in our understanding of their performance on the complex irregularities intrinsic to ICU data. This motivates the central questions of our work: \emph{To what extent can current LLMs model the complex irregularities in ICU time series data, and do they offer practical advantages over existing irregular-specific lightweight models?}}


This paper aims to bridge the gap by investigating LLMs for irregular time series, focusing on their effectiveness, limitations, and computational trade-offs relative to time series models without LLMs. We highlight two key components underlying LLM-based time series modeling:
(1) the \emph{time series encoder}, which transforms raw or preprocessed time series signals into high-level representations; and
(2) the \emph{multimodal alignment strategy}, which aligns temporal and textual representations suitable for LLM processing.
\begin{figure}[t]
    \centering
    \includegraphics[trim={0.1cm 0 0 0}, clip, width=0.95\linewidth]{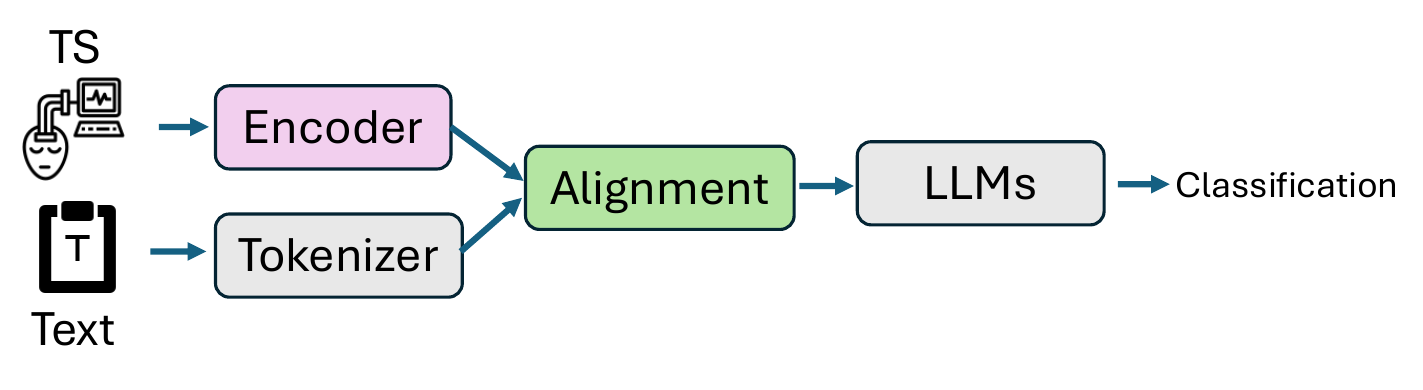}
    \vspace{-15pt}
    \caption{LLM architectures for time series (TS) modeling.}
    \label{fig:LLM_based_model}
    \vspace{-15pt}
\end{figure}

To systematically rethink the applicability and limitations of these key components in processing irregular time series, we propose the following research questions:

\noindent \textbf{RQ1.} To what extent can LLMs developed for regular time series effectively generalize to irregular ICU healthcare data? 

\noindent \textbf{RQ2.} Which components of time series LLMs have the greatest impact on modeling irregular time series? 

\noindent \textbf{RQ3.} What are the trade-offs between the computational overhead introduced by LLMs and the performance gains they offer for irregular time series modeling, compared to traditional methods? 

\noindent \textbf{RQ4.} How effective are time series LLMs at few-shot learning for irregular time series?

To address these questions, we conduct a comprehensive empirical study using four state-of-the-art (SOTA) LLM-based methods, specifically, Time-LLM \cite{time_llm}, $S^2$IP \cite{S2IP_llm}, CALF \cite{calf}, and FSCA \cite{FSCA}, in comparison with traditional supervised and self-supervised baselines on ICU mortality prediction task using two benchmark irregular ICU datasets.
To the best of our knowledge, this is the first systematic effort to assess the utility of LLMs for irregular time series classification in critical care. Our work offers actionable insights for designing more effective and computationally viable LLM-based time series models for clinical applications. 

\vspace{-5pt}
\section{Related Work}
\vspace{-5pt}
\begin{figure*}[t]
  \centering
  \includegraphics[width=0.9\textwidth]{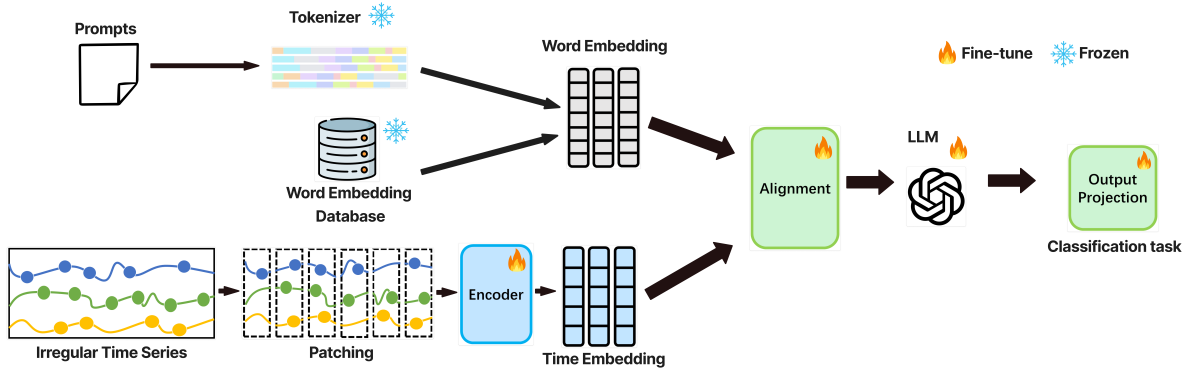}
  \vspace{-10pt}
  \caption{\textbf{Workflow} of LLMs. Irregular ICU time series is encoded and aligned with textual embeddings for classification.}
  \vspace{-10pt}
  \label{fig:workflow}
\end{figure*}


\textbf{Traditional irregular time series modeling.}
Early approaches addressed irregularity through imputation or aggregation, but these strategies often discarded information and failed to capture fine-grained temporal dynamics \cite{white2011multiple, che2018recurrent}. Recent efforts have developed \textit{encoders} that explicitly handle irregularity. Specifically, Latent ODEs \cite{latent_ODE} model continuous-time dynamics via ordinary differential equations, while attention-based models such as mTAND \cite{mTAN} uses multi-time attention and Warpformer \cite{warpformer} introduce time-warping layers and doubly self-attentive blocks. Despite their effectiveness, these methods remain fully supervised and are validated on limited datasets, restricting scalability and generalization.

\noindent\textbf{LLMs for time series modeling.}
LLMs have demonstrated strong generalization and reasoning capabilities across a wide range of NLP and multimodal tasks, motivating their application to time series.
Early works leveraged prompt-based methods, treating numeric time series as text to process \cite{xue2023promptcast}, but these approaches neglected the modality gap between textual and numerical data, limiting their robustness \cite{calf}. Recent research addresses this gap by introducing explicit alignment techniques. Approaches such as Time-LLM reprograms time series into to LLM embedding spaces, while $S^2$IP \cite{S2IP_llm} leverages semantic anchors for cross-modal alignment. CALF \cite{calf} extends this by enforcing feature-level and output-level consistency, and FSCA \cite{FSCA} introduces structural and logical alignment using graph neural networks. However, existing studies have concentrated primarily on regularly sampled datasets, overlooking irregular time series data that frequently arise in healthcare. Our work bridges this gap by investigating the effectiveness of time series encoder structures and alignment methods in handling irregular data.

\vspace{-5pt}
\section{Experimental Setup}
\vspace{-5pt}

We design a comprehensive empirical testbed with four SOTA LLM-based methods to examine the role of \textit{time series encoders} and \textit{multimodal alignment strategies} in TSM, as shown in Figure \ref{fig:workflow}. In this framework, irregular time series are segmented into fixed-size patches and encoded into temporal embeddings. In parallel, auxiliary textual embeddings are obtained either by tokenizing and encoding clinical text or by leveraging pretrained word token embeddings. An alignment module then fuses temporal and textual representations, which are fed into the LLM backbone. Finally, the output is passed through a classification head to support downstream tasks.


\noindent\textbf{Baselines.}
We conduct a comprehensive empirical testbed using four SOTA LLM-based models (\textit{Time-LLM} \cite{time_llm}, \textit{$S^2$IP} \cite{S2IP_llm}, \textit{CALF} \cite{calf} and \textit{FSCA} \cite{FSCA}), two SOTA self-supervised foundation models (\textit{MOMENT} \cite{moment} and \textit{UniTS} \cite{UniTS}), and SOTA supervised models for irregular time series (\textit{mTAND} \cite{mTAN} and \textit{Warpformer} \cite{warpformer}).  

\noindent\textbf{Encoder and Alignment.} For the encoder, we benchmark three architectures identified in our baseline LLMs for TSM, and include an additional time-aware encoder (mTAND)~\cite{mTAN} specifically designed to handle irregularity. For the alignment strategy, we investigate four distinct approaches from the baseline models, ranging from traditional cross-modal attention~\cite{calf} to more advanced graph-based and semantic prompting methods~\cite{S2IP_llm}. A detailed discussion of these components is provided in Section \ref{sec:4.2}.

\noindent\textbf{Datasets.}
We evaluate on two large real-world ICU datasets and a semi-synthetic irregular ECG benchmark, enabling direct comparison between regular and irregular time series.
\begin{itemize}[leftmargin=11pt]
    \item  \textbf{PhysioNet 2012:} This dataset comprises ICU records from over 12,000 patients with 41 physiological variables and a binary label indicating in-hospital mortality \cite{physionet2012}. Following the preprocessing procedure \cite{raindrop}, we remove 12 problematic samples. 
    The task is to predict in-hospital mortality.

 \item\textbf{MIMIC-III:} The MIMIC-III database contains ICU electronic health records. We follow the existing preprocessing pipeline \cite{bilovs2021neural, ir_gnn}, extracting time-varying measurements for 96 physiological variables across 24,681 patients. 
The task is in-hospital mortality classification.

\item \textbf{Semi-synthetic irregular ECG:} The MIT-BIH Atrial Fibrillation dataset \cite{ECG_data} contains two-channel ECG recordings from 25 subjects across four arrhythmia classes. We follow the previous study \cite{unsup_ECG} to preprocess the dataset. The dataset is originally regular. To enable a fair comparison of models on the same types of regular and irregular time series, we randomly drop data points to simulate irregularity.

\end{itemize}


\begin{table}[t!]
  \centering
  \vspace{-1em}
  \caption{Dataset Statistics}
  \setlength{\tabcolsep}{3pt}
  \label{tab:dataset_statistics}
  \resizebox{\columnwidth}{!}{%
  \begin{tabular}{@{}lccccc@{}}
    \toprule
    \textbf{Dataset}      & \textbf{Samples} & \textbf{Variables} & \textbf{Classes} & \textbf{Missing Ratio} \\
    \midrule
    PhysioNet 2012 & 11,988 & 41 & 2   & 85.7\%\\
    MIMIC-III & 24,681 & 96  & 2   & 96.7\% \\
    MIT-BIH & 59,850 & 2 & 4  & 0\% - 90\% \\
    \bottomrule
  \end{tabular}
  }
  \vspace{-1em} 
\end{table}


\noindent\textbf{Training and evaluation setup.}
Datasets are split into training, validation, and test sets as follows: 60\%/20\%/20\% for the ECG dataset, 80\%/10\%/10\% for the PhysioNet 2012 dataset \cite{raindrop}, and 60\%/20\%/20\% for the MIMIC-III dataset \cite{ir_gnn}. All models are trained on an NVIDIA A100 80GB GPU, except for Time-LLM, which requires two A100 GPUs due to high memory consumption.
To ensure reproducibility and stability, each experiment is repeated three times and we report the mean and standard deviation of evaluation metrics. 
For PhysioNet 2012 and MIMIC-III datasets, we report the performance in terms of the Area Under the Precision-Recall Curve (AUPRC) and the Area Under the Receiver Operating Characteristic Curve (AUROC) since they are highly class imbalanced. For semi-synthetic ECG dataset, we report AUPRC and accuracy following existing work \cite{unsup_ECG}.

\begin{table*}[t]
  \centering
  \caption{Comparison of \textbf{encoder and alignment strategies} for LLMs on irregular ICU time series. The best result for each metric is shown in \textbf{bold}, and the second-best is \underline{underlined}.}
  \resizebox{\textwidth}{!}{%
    \begin{tabular}{@{}lcccc l cccc@{}}
      \toprule
      & \multicolumn{4}{c}{\textbf{Encoder}} & & \multicolumn{4}{c}{\textbf{Alignment}} \\
      \cmidrule(lr){2-5} \cmidrule(lr){7-10}
      & \multicolumn{2}{c}{\textbf{PhysioNet 2012}} & \multicolumn{2}{c}{\textbf{MIMIC-III}}
      & 
      & \multicolumn{2}{c}{\textbf{PhysioNet 2012}} & \multicolumn{2}{c}{\textbf{MIMIC-III}} \\
      \cmidrule(lr){2-3}\cmidrule(lr){4-5}\cmidrule(lr){7-8}\cmidrule(lr){9-10}
      \textbf{Methods} & \textbf{AUPRC} & \textbf{AUROC} 
                       & \textbf{AUPRC} & \textbf{AUROC} 
      & \textbf{Methods} 
                       & \textbf{AUPRC} & \textbf{AUROC} 
                       & \textbf{AUPRC} & \textbf{AUROC} \\
      \midrule
      1DCNN & 32.9$\pm$3.1 & 76.0$\pm$1.7 & 18.6$\pm$1.1 & 66.1$\pm$1.1 &
      Reprogramming & 51.1$\pm$2.5 & 84.9$\pm$1.1 & 23.4$\pm$0.2 & 74.2$\pm$0.5 \\
      Decomposition & 21.8$\pm$2.3 & 64.9$\pm$1.4 & 26.1$\pm$1.3 & 74.1$\pm$2.1 &
      \textbf{$\mathbf{S^2}$IP }    & \textbf{54.0$\pm$2.5} & \textbf{85.8$\pm$1.2} & \textbf{41.8$\pm$0.8} & \textbf{83.4$\pm$0.5} \\
      Transformer & 35.7$\pm$3.4 & 77.3$\pm$2.4 & 29.0$\pm$1.1 & 76.7$\pm$1.1 &
      Cross-modal   & 53.5$\pm$2.7 & 85.8$\pm$1.1 & 38.7$\pm$1.4 & 82.0$\pm$0.8 \\
      \textbf{mTAND} & \textbf{53.5$\pm$2.7} & \textbf{85.8$\pm$1.1} & \textbf{38.7$\pm$1.4} & \textbf{82.0$\pm$0.8} &
      Context Alignment     & 47.4$\pm$1.8 & 82.4$\pm$2.1 & 19.0$\pm$2.5 & 68.7$\pm$3.6 \\
      \bottomrule
    \end{tabular}%
    }
    \label{tab:encoder_alignment_comparison}
    \vspace{-1em}
\end{table*}
\vspace{-5pt}
\section{Results and Findings}
\vspace{-5pt}
\label{results_findings}

\subsection{RQ1. 
To what extent can LLMs developed for regular time series effectively generalize to irregular ICU healthcare data?} 
\textbf{Setup.}
To enable fair comparison, the ideal scenario is to use both regular and irregular versions of the same data types. Since no such real-world data exist, we utilized a semi-synthetic ECG dataset by randomly dropping data points to simulate sampling gaps, with missing ratio from 0\% to 90\%. This design allows systematic evaluation of robustness of existing LLMs under different levels of irregularity.


\begin{figure}[t!]
    \centering
    \includegraphics[width=0.9\linewidth]
    {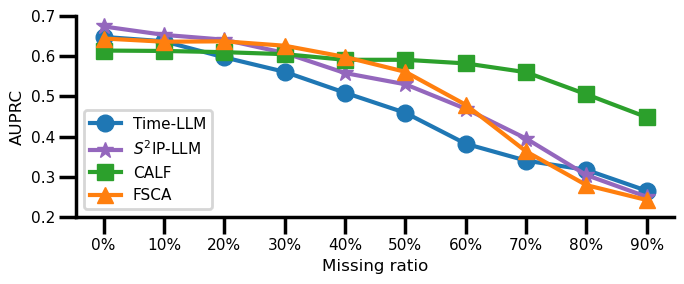}
    \vspace{-1em}
    \caption{\textbf{AUPRC of LLMs on the semi-synthetic irregular ECG dataset} as the proportion of missing values increases.}
    \label{fig:auprc_vs_dropped_percentage}
    \vspace{-1em} 
\end{figure}

\noindent\textbf{Findings.}
Figures \ref{fig:auprc_vs_dropped_percentage} shows the AUPRC across missing ratios. For the regular time series (\textit{i.e.}, 0\% missing rate), all LLM-based methods achieved strong performance, with AUPRC above 0.6. However, with the missing ratio increasing from 10\% to 90\%, Time-LLM, $S^2$IP and FSCA drop by 38\%, 42\%, and 40\%, respectively, reflecting poor robustness. This poor performance stems from their patching mechanism, where higher missing rates leave patches with too few valid data for effective contextual learning. In contrast, CALF exhibits a more gradual performance drop, as it does not apply patching and instead captures global representation over the entire sequence rather than local segments using attention. As a result, even when partial observations are missing, the model can still exploit coarse-grained temporal correlations across the full sequence. Therefore, such design may offer a more resilient representation strategy when facing irregularity.

\vspace{-5pt}
\subsection{RQ2. Which components of LLMs for time series have the greatest impact on irregular time series modeling? Is it \textit{encoder} or \textit{alignment} between textual and temporal representations?
} \label{sec:4.2}

\noindent\textbf{Setup.}
As demonstrated in \textbf{RQ1}, directly applying existing LLM-based models to irregular time series leads to significant performance degradation. To understand and mitigate this issue, we investigate two key components in LLM-based time series modeling: (1) \textbf{\textit{encoder}}, 
and (2) \textbf{\textit{alignment strategy}}.

Based on \textbf{RQ1}, we selected the best-performing CALF as the representative framework for analysis. First, to assess encoder effects, we replace the default encoder of CALF (a single-layer Transformer with 12 heads) with the following alternatives from existing approaches, while keeping the other components unchanged: (i) \textbf{1DCNN} (Time-LLM and FSCA): A one-dimensional convolutional neural network;
(ii) \textbf{Decomposition Encoder} ($S^2$IP): Decomposes the time series into trend, seasonality, and residual components, then maps these into a high-dimensional vector;
and (iii) \textbf{mTAND}: Specifically designed for irregular time series, utilizing continuous-time encoding and multi-timescale attention. We included this irregularity-aware encoder into LLM framework to directly compare its effectiveness against the conventional encoders (e.g., vanilla Transformer) used in existing LLMs designed for regular time series.

Next, we fix the best-performing encoder from the previous step and compare four alignment strategies:
(i) \textbf{Reprogramming }(Time-LLM): Applies a cross-attention module to transform time series patches into language-like tokens, guided by prefix prompts with dataset-level context; 
(ii) \textbf{Semantic Space Informed Prompting} ($S^2$IP): Aligns time series features with pretrained LLM embeddings, retrieving top-K semantic anchors as prefix prompts;
(iii) \textbf{Cross-modal Fine-tuning} (CALF): Uses dual-branch attention and consistency losses (feature- and output-level) to align time series and textual modalities;
and (iv) \textbf{Graph-based Context Alignment} (FSCA): Represents time series as hierarchical graph nodes and edges to capture structural and logical relations, enabling LLMs to contextualize inputs with linguistic priors.

\noindent\textbf{Findings:}
\noindent \textit{\underline{Encoder:}} As shown in Table \ref{tab:encoder_alignment_comparison}, mTAND outperforms other encoders, owing to its explicitly incorporating of irregular-aware time embeddings. 
By contrast, since the 1D CNN, decomposition encoder, and Transformer encoder are originally designed for regularly sampled inputs, they assume uniform temporal spacing between observations. Therefore, they introduce misleading temporal biases when applied to irregular data. 
This finding emphasizes the importance of using encoder tailored to the temporal characteristics of irregular time series.


\begin{table*}[t]
\centering
\caption{\textbf{Computational cost}, measured as training time (hours:minutes:seconds), for classifying irregular ICU time series.}
\begin{tabular}{@{}lcccccc@{}}
\toprule
\textbf{Model} 
  & \multicolumn{3}{c}{\textbf{PhysioNet 2012}} 
  & \multicolumn{3}{c}{\textbf{MIMIC-III}} \\
\cmidrule(lr){2-4}\cmidrule(lr){5-7}
 & \textbf{AUPRC} & \textbf{AUROC} & \textbf{Training Time}
 & \textbf{AUPRC} & \textbf{AUROC} & \textbf{Training Time} \\
\midrule
\multicolumn{7}{@{}l@{}}{\textbf{Supervised}} \\
mTAND         & 51.7±0.5          & 85.2±0.6          & 00:07:18 
             & 34.6±1.2          & 80.3±0.7          & 00:08:51 \\
Warpformer   & \textbf{54.8±2.0} & \textbf{86.6±1.0} & 00:13:41 
             & \underline{38.2±1.1} & \underline{82.7±1.1} & 01:42:06 \\
\midrule
\multicolumn{7}{@{}l@{}}{\textbf{Self-supervised}} \\
MOMENT       & 30.8±2.3          & 73.0±0.9          & 09:42:07 
             & 25.1±1.7          & 74.7±0.1          & 40:16:54 \\
UniTS        & 26.3±1.3          & 70.1±0.8          & 01:08:42 
             & 25.2±0.9          & 75.0±2.1          & 04:15:35 \\
\midrule
\multicolumn{7}{@{}l@{}}{\textbf{LLM-based}} \\
Time-LLM     & 29.0±7.7          & 72.2±4.5          & 02:33:02 
             & 25.7±3.0          & 75.2±2.2          & 09:26:22 \\
$S^2$IP   & 38.5±1.4          & 78.8±1.8          & 07:33:35 
             & 30.3±1.1          & 77.7±0.4          & 23:15:38 \\
CALF         & 32.1±2.8          & 74.6±1.6          & 00:11:45 
             & 18.4±0.9          & 65.9±0.7          & 00:26:07 \\
FSCA         & 32.2±6.5          & 74.7±3.5          & 01:05:03 
             & 33.7±3.5          & 79.0±1.8          & 02:05:31 \\ \midrule
mTAND\(+\)$S^2$IP
             & \underline{54.0±2.5} & \underline{85.8±1.2} & 02:15:59 
             & \textbf{41.8±0.8}    & \textbf{83.4±0.5}    & 24:38:50 \\
\bottomrule
\end{tabular}%
\label{tab:performance_P12_MIMIC}
\vspace{-10pt}
\end{table*}

\noindent \textit{\underline{Alignment:}} As shown in Table~\ref{tab:encoder_alignment_comparison}, with mTAND encoder fixed, among the alignment techniques, the Semantic Space Informed Prompting used in $S^2$IP approach achieves the best performance, benefiting from semantic anchor retrieval that aligns time series with relevant pretrained word embeddings. In contrast, cross-attention-based approaches distribute attention across multiple keys, which may lead to overly diffuse or redundant representations that lack semantic focus. The graph-based alignment strategy adopted by FSCA performs poorly, suggesting that its graph structure may not be well-suited for modeling sparsely sampled or noisy clinical time series when the graph node representation loses reliability.

\noindent \textit{\underline{Overall Insight:}}
The choice of encoder has a larger impact than alignment: on PhysioNet 2012, the AUPRC gap between the best and worst encoders is 31.7\%, versus 6.6\% for alignment strategies. While alignment strategies can improve performance, their effect is less than encoder, indicating the need for further development in customized alignment with irregular time series.


\begin{table}[t]
\centering
\vspace{-5pt}
\caption{\textbf{Few-shot classification} on irregular ICU time series using various LLM configurations for TSM.}
\scriptsize
\resizebox{\columnwidth}{!}{%
\begin{tabular}{@{}lcccc@{}}
\toprule
\textbf{Model} & \multicolumn{2}{c}{\textbf{PhysioNet 2012}} & \multicolumn{2}{c}{\textbf{MIMIC-III}} \\
\cmidrule(lr){2-3} \cmidrule(lr){4-5}
& \textbf{AUPRC} & \textbf{AUROC} & \textbf{AUPRC} & \textbf{AUROC} \\
\midrule
Warpformer  & \textbf{46.0$\pm$2.7} & \textbf{82.2$\pm$1.0} & \textbf{30.7$\pm$1.9} & \textbf{78.0$\pm$1.0} \\
$S^2$IP     & 28.3$\pm$2.3 & 72.3$\pm$1.0 & 22.6$\pm$1.5 & 71.3$\pm$1.8 \\
mTAND\(+\)$S^2$IP   & 36.2$\pm$9.8 & 73.3$\pm$6.6 & 22.2$\pm$1.2 & 71.1$\pm$0.5 \\
\bottomrule
\end{tabular}
}
\vspace{-10pt}
\label{tab:few_shot_results}
\end{table}

\vspace{-5pt}
\subsection{RQ3. What are trade-offs between computational overhead introduced by LLMs and the gains they offer for irregular time series modeling, compared to traditional methods? 
}
\noindent\textbf{Setup.}
The use of LLMs undoubtedly introduces additional computational overhead \cite{kaplan2020scaling}. 
To assess whether this cost is justified, we compare SOTA LLM-based models with SOTA self-supervised models, and SOTA supervised models tailored for irregular time series, reporting both classification performance and training time as a proxy for computational efficiency. We prioritize training time over inference, as inference is generally comparable across LLM approaches and typically negligible relative to training.
Based on \textbf{RQ2}, we also evaluate a hybrid model that combines the best encoder (mTAND) with the best alignment method ($S^2$IP).

\noindent\textbf{Findings.} 
Table \ref{tab:performance_P12_MIMIC} shows that on the PhysioNet 2012 dataset, among the four SOTA LLM-based models, $S^2$IP achieves the highest performance, likely attributable to its superior multimodal alignment strategy. Notably, the combined approach of (mTAND\(+\)$S^2$IP) surpasses all four LLM-based models, with relative improvements of 15.5\% in AUPRC and 7\% in AUROC, highlighting the critical importance of advanced encoder designs for learning from irregular time series. 
However, it still lags behind Warpformer, which achieves the highest performance despite not leveraging LLMs. In terms of computational efficiency, LLM-based methods train faster than self-supervised foundation models but are far less efficient than supervised approaches: the training time for (mTAND\(+\)$S^2$IP) is 9.94$\times$ longer than that of Warpformer.
On the MIMIC-III dataset, (mTAND\(+\)$S^2$IP) achieves the highest performance, yielding relative improvements of 3.6\% in AUPRC and 0.7\% in AUROC compared to Warpformer. However, these modest gains in predictive performance come at a significant computational cost: the training time for (mTAND\(+\)$S^2$IP) is approximately 14 times longer than that of Warpformer, offering extremely limited practical benefit.
\emph{\underline{Overall Insight}}: The increased computational overhead associated with LLMs does not yield commensurate improvements in classification performance, suggesting that the incorporation of LLMs may not be warranted for irregular time series tasks.


\subsection{\textbf{RQ4. How effective are time series LLMs at the few-shot learning for irregular time series?}}

\noindent\textbf{Setup.}
LLMs possess strong few-shot learning capabilities \cite{brown2020language, achiam2023gpt}. To assess whether LLM-based models demonstrate comparable few-shot learning ability in irregular time series classification, we follow the experimental protocols of existing work \cite{calf, one_fits_all}. We employ only 10\% of the training data as few-shot examples, while evaluating on the same test dataset. Based on \textbf{RQ3}, we compare the three best-performing models: mTAND\(+\)$S^2$IP, $S^2$IP, and Warpformer.

\noindent\textbf{Findings.} As shown in Table~\ref{tab:few_shot_results}, Warpformer model outperforms the LLM-based approaches of (mTAND\(+\)$S^2$IP) and $S^2$IP, achieving the highest AUPRC and AUROC scores. 
In contrast, both (mTAND\(+\)$S^2$IP) and $S^2$IP perform worse, indicating that the integration of LLMs does not confer an advantage in the few-shot learning setting for irregular time series classification.  One possible explanation is that LLMs, pretrained on large text corpora and containing billions of parameters, may be less effective at learning from limited irregular time series data. In contrast, domain-specific models like Warpformer, with smaller and more efficient architectures, generalize better under sparse supervision. \emph{This suggests that a current limitation in adapting LLMs to irregular time series tasks, particularly under data-scarce conditions.}



\section{Conclusion}
This study systematically investigated LLM-based approaches for irregular time series classification tasks in critical care settings and found that encoder architecture is a key determinant of performance. Additionally, incorporating LLMs offers slight improvements at significant computational cost and does not resolve generalization issues. These results highlight the need for more efficient architecture that balances performance improvements and computational costs to deliver reliable insights in resource-constrained clinical settings.

\begin{small}
\bibliographystyle{IEEEbib}
\bibliography{strings,refs}
\end{small}

\end{document}